\documentclass[preprint,12pt,authoryear]{elsarticle}
\newif\ifreviewversion
\reviewversionfalse
\newcommand{\versionlabel}{Preprint draft, 6 November 2026}
\usepackage[T1]{fontenc}
\usepackage[utf8]{inputenc}
\usepackage{lmodern,amsmath,amssymb,booktabs,array,graphicx,microtype}
\usepackage[hidelinks]{hyperref}
\newcommand{\papertitle}{Efficient 3D plant segmentation through multiscale graph learning and optimized geometric processing}

\biboptions{sort&compress}
\makeatletter
\def\ps@pprintTitle{\let\@oddhead\@empty\let\@evenhead\@empty
\def\@oddfoot{\hfil\footnotesize\versionlabel\hfil}
\let\@evenfoot\@oddfoot}
\makeatother

\begin{document}
\begin{frontmatter}
\title{\papertitle}
\author[guadalajara]{Jesus Ivan Ruiz-Martinez}
\ead{jesus.ruiz@cinvestav.mx}
\author[guadalajara]{Andr\'es Mendez-Vazquez\texorpdfstring{\corref{correspondence}}{}}
\ead{andres.mendez@cinvestav.mx}
\author[tamaulipas]{Eduardo Rodriguez-Tello\texorpdfstring{\corref{correspondence}}{}}
\ead{ertello@cinvestav.mx}
\cortext[correspondence]{Corresponding authors.}
\affiliation[guadalajara]{organization={Cinvestav Unidad Guadalajara},
addressline={Av. del Bosque 1145},city={Zapopan},postcode={45017},state={Jalisco},country={Mexico}}
\affiliation[tamaulipas]{organization={Cinvestav Unidad Tamaulipas},
addressline={Km. 5.5 Carretera Victoria-Soto La Marina},city={Victoria},postcode={87130},state={Tamaulipas},country={Mexico}}

\begin{abstract}
Plant organ segmentation on compact platforms requires accurate predictions and affordable geometric preparation. We evaluate a multiscale graph attention model combining local shape descriptors, two neighborhood scales and residual organ correction in approximately 78 thousand parameters. Predictions and metrics cover 1,024 sampled points per cloud, without dense reconstruction. TomatoWUR inputs are annotation-prefiltered to exclude supports and ignored points. Broad local comparisons identify PointNeXt as the strongest tested external accuracy reference. A nested four-variant ablation crosses geometric information with the combined multiscale configuration across seven outer groups and three seeds. The full model achieves the highest mean intersection over union among these variants, 86.36\%, although uncertainty and boundary errors qualify component claims. Optimized preparation preserves full cloud principal component analysis, sampling and graph definitions without changing learned weights. All 1,180 desktop and 3,540 repeated ARM comparisons preserve exact logits and predictions within their respective platforms. Multiscale pipeline latency decreases by 70.69\% on PlantNet and 56.08\% on TomatoWUR on desktop, and by 63.46\% and 54.22\% on Raspberry Pi 5. On Pi, optimized Multiscale averages 277.73 milliseconds versus PointNeXt's 342.47 on PlantNet, but 1,196.28 versus 458.51 on TomatoWUR. PointNeXt also receives optimized sampling and remains faster on desktop and at PlantNet's Pi p95. Three repeated ARM cycles characterize inference timing, not new training seeds or biological samples. These results establish reduced geometric overhead and physical inference feasibility, without demonstrating energy efficiency or readiness for autonomous flight.

\end{abstract}
\end{frontmatter}
\section{Introduction}
Precision agriculture increasingly uses spatial observations to understand variation within a field and guide management. Small aerial systems offer flexible observation schedules and detailed imagery, with a substantial body of work on agricultural monitoring \citep{zhang2012}. Recent systems also investigate inference close to the sensor, including weed detection from aerial imagery \citep{agriuav}. For agricultural regions where repeated field inspection and timely local decisions matter, compact perception models could complement these tools by reducing dependence on a workstation or a continuous connection. This is an engineering motivation rather than a measured socioeconomic outcome of the present study.

Three dimensional plant segmentation addresses a different level of detail from canopy mapping. Separating stems, leaves and substrate can support structural phenotyping, yet thin stems and attachment regions remain difficult. Public datasets illustrate substantial changes in acquisition, organ definitions and plant morphology across species \citep{pheno4d,psegnet,tomatowur}. A model that performs well on maize cannot therefore be assumed to work equally well on tomato or other crops. An eventual drone application would also require sufficient geometric resolution, reliable reconstruction and a suitable payload. Here, plant clouds are already available; acquisition and flight are outside the measured system.

The evaluation begins with a broad comparison rather than a single competing model. The recorded benchmarks include PointNet++, DGCNN, Point Transformer, RandLA-Net, PSegNet and PointNeXt on maize and other plant collections. Additional experiments on PlantNet and TomatoWUR include Point Transformer V3 and PointCT adapters. Within these local recipes, PointNeXt provides the strongest external accuracy reference across the available collections. Its performance motivates the subsequent focused comparison with Multiscale; it is selected from the observed benchmark, not assumed in advance to be the only relevant alternative. The complete baseline tables remain part of the main results, including cases where our context variant or Multiscale itself has a higher mean. This local ranking does not establish superiority over untested recent plant segmentation methods.

The engineering problem is not simply to reduce the number of learned weights. Geometric descriptors, point selection and neighborhood construction can dominate a compact network's runtime. Moving an expensive operation outside the neural network does not remove it from the deployed pipeline. Hardware measurements are consequently more informative than parameter count alone \citep{netadapt}. This matters for Multiscale EdgeGAT: its small network uses local geometric descriptors from principal component analysis (PCA), and preparing those descriptors on a dense cloud can cost more than inference.

After establishing this accuracy context, we investigate whether that geometric information can be retained at lower implementation cost. The proposed preparation combines an exact compiled implementation of farthest point sampling with parallel PCA and neighbor queries. It changes execution, not learned weights, descriptor definitions or model topology. The experimental identifier P4 is retained only to connect this version to the frozen records. Throughout the paper the model is called Multiscale EdgeGAT or Multiscale; development identifiers are confined to the reproducibility notes.

The contribution is a compact plant segmentation design evaluated at both model and pipeline level. We establish its accuracy relative to the tested baseline families, then examine geometric information and the multiscale configuration through a completed nested factorial ablation. A paired implementation study shows how to retain the frozen representation at lower preparation cost. Finally, stagewise desktop measurements and three repeated Raspberry Pi cycles identify workload and platform dependent latency tradeoffs. The design combines established edge learning and attention mechanisms; it does not introduce a new PCA estimator or claim a new general message passing operator. Its value must therefore be assessed through the controlled comparisons and the measured implementation gains.

\section{Related work}
\subsection{Point processing and local geometry}
PointNet uses shared point transformations and symmetric aggregation \citep{pointnet}; PointNet++ introduces hierarchical neighborhood learning \citep{pointnetpp}. EdgeConv forms messages from a point and its neighbors and is central to Dynamic Graph CNN \citep{dgcnn}. Graph attention learns how to weight neighboring features \citep{gat}. Multiscale EdgeGAT combines these established mechanisms on coordinate neighborhoods, rather than dynamically rebuilding graphs from learned features as in DGCNN. GLA-Net already combines graph convolution, attention and local/global context; its neighborhood module uses ordinary and dilated kNN with generalized mean pooling \citep{glanet}. Our nested coordinate neighborhoods of sizes 8 and 16 and maximum aggregation are different implementation choices, not evidence that combining graphs and attention is new.

Efficient segmentation also depends on sampling, aggregation and training. RandLA-Net uses random sampling and local aggregation \citep{randla}, while PointNeXt revisits PointNet++ with improved training and scaling \citep{pointnext}. Point Transformer and Point Transformer V3 provide attention based alternatives \citep{pointtransformer,ptv3}. More recently, LitePT places convolution in early, higher resolution stages and attention in deeper, lower resolution stages \citep{litept}. Thus PointNeXt is a strong reference among our tested recipes, not a complete representation of current efficient point cloud methods. The local implementations and adapters in our experiments are not official benchmark reproductions, and their ranking cannot be assigned to entire model families.

\subsection{Plant architectures and prior dissemination}
PSegNet combines EdgeConv, attention and fusion at two granularities for plant semantic and leaf instance segmentation \citep{psegnet}. Its double neighborhoods are defined in coordinate and learned feature spaces; this differs from two sizes within our coordinate space. Its joint semantic/instance task and voxelized sampling also differ from our sampled organ classification. PlantNet similarly combines dynamic graph feature extraction with semantic/instance fusion and edge preserving sampling \citep{plantnet}. PlantNet denotes that published network, whereas our tables use the same name as a local collection identifier; they do not report an evaluation of the PlantNet network.

Recent plant methods make the architectural overlap more specific. GCASSN combines local graph convolution and global self-attention from XYZ inputs with an explicit efficiency motivation \citep{gcassn}. Its reported epoch selection uses test mIoU, so its published scores cannot be interpreted as paired comparisons with our inner/outer AO protocol. GSAN uses graph self-attention, fusion across scales and a semantic discriminative objective for stem--leaf separation \citep{gsan}. ELGCot3D combines EdgeConv, local attention and graph convolution for cotton and other crops, reporting 0.86 million parameters \citep{elgcot3d}. Its task and recipe differ from ours; a smaller parameter count alone would establish neither higher accuracy nor lower latency. PlantEFRSegnet uses PCA informed edge identification, edge preserving sampling, feature repair and point attention \citep{plantefrsegnet}. This changes the selected representation, whereas P4 retains the original sampling and descriptors. These works preclude a general novelty claim for lightweight plant graph learning, explicit geometry or multiscale attention.

The preprint of \citet{edgegat2025} is an earlier version of the present manuscript. It already describes XYZ plus ten PCA descriptors, coordinate kNN, two EdgeConv blocks and graph attention for stem, leaf and soil, with 1,024 input points and $k=16$. These shared components are not presented here as new. The present version adds the evaluated K8/K16 configuration with organ logit refinement, broader local comparisons, separate inner checkpoint selection and outer AO evaluation, the completed factorial analysis, and preparation and physical ARM evidence. Table~\ref{tab:antecedent} distinguishes this added scope from the earlier preprint. The preprint and this manuscript are not independent studies, and their scores are not used as a matched before/after accuracy comparison.

\begin{table}[tbp]\centering\small
\caption{Scope relative to the earlier preprint version, arXiv:2507.00182v2. Differences describe reported methods and evidence, not a matched performance comparison.}\label{tab:antecedent}
\begin{tabular}{@{}>{\raggedright\arraybackslash}p{0.20\linewidth}>{\raggedright\arraybackslash}p{0.34\linewidth}>{\raggedright\arraybackslash}p{0.39\linewidth}@{}}\toprule
Aspect & Earlier preprint \citep{edgegat2025} & Present study\\\midrule
Shared core & XYZ, ten PCA channels, coordinate kNN, EdgeConv and GAT & Same broad representation and established operator families\\
Configuration & Neighborhood size 16 & K8/K16 backbone; separate organ logit correction\\
Evaluation & Fold validation on maize & Historical collection comparisons; nested AO factorial with explicit provenance limits\\
Implementation & No paired P4 or ARM study reported & Frozen prediction checks and stage timings on desktop and Pi; three repeated ARM cycles on PlantNet/TomatoWUR\\\bottomrule
\end{tabular}
\end{table}

\subsection{Collections, task definitions and output resolution}
Pheno4D provides repeated observations of maize and tomato \citep{pheno4d}. Single Maize is associated with a deformable point cloud framework for maize organ segmentation \citep{singlemaize}. PLANesT-3D covers pepper, rose and red currant \citep{planest}; the local PlantNet collection contains tobacco, tomato and sorghum associated with the PlantNet/PSegNet research line \citep{plantnet,psegnet}. TomatoWUR supplies semantic labels together with skeleton and trait references \citep{tomatowur}. Organ taxonomies are not interchangeable. Merging main and side stems and excluding support poles creates a binary task distinct from TomatoWUR's original taxonomy.

Our citations establish attribution and context; local manifests establish the files, labels and groups actually evaluated. Repeated captures and augmented samples are not independent plants. Historical validation scores are separated from nested evaluation because a fold used to select an epoch is not an independent outer test. Resolution is another distinction: OmniPlantSeg's subset construction enables predictions across the full input cloud \citep{omniplantseg}. Our evaluation predicts and scores only 1,024 sampled points per cloud, without label propagation or full resolution reconstruction. Full cloud PCA support must not be mistaken for dense output coverage.

\subsection{Efficiency and embedded evaluation}
EPoNET studies quantization and feature refinement for efficient large-scale multispectral LiDAR perception \citep{eponet}; P4 instead optimizes execution of a fixed floating point representation. Their published acceleration factors do not share our hardware, tasks or timer boundaries. Efficient tree segmentation has also been studied using RGB-D inputs, a simplified backbone and contour coefficients \citep{treesembedded}. That work's reported 53 frames per second uses an RTX 2060 Super and concerns tree contours, not Raspberry Pi organ classification from 3D points.

NetAdapt and MCUNet connect architecture to actual platform resources \citep{netadapt,mcunet}. Visual navigation for nano drones demonstrates onboard inference under specific sensing and power constraints \citep{palossi}, but those measurements cannot be transferred to our workload. Compact weights do not imply low process memory, and successful execution establishes neither energy efficiency nor safe flight. We use Raspberry Pi 5 for physical ARM inference measurements. The unresolved comparison is whether our complete recipe offers a useful accuracy/cost tradeoff against recent plant and efficient point methods under common inputs, splits and hardware; the present tables establish that tradeoff only for the recipes actually evaluated.

\section{Model and optimized geometric preparation}
\subsection{Multiscale EdgeGAT}
\begin{figure}[p]\centering
\makebox[\linewidth][c]{\includegraphics[width=190mm]{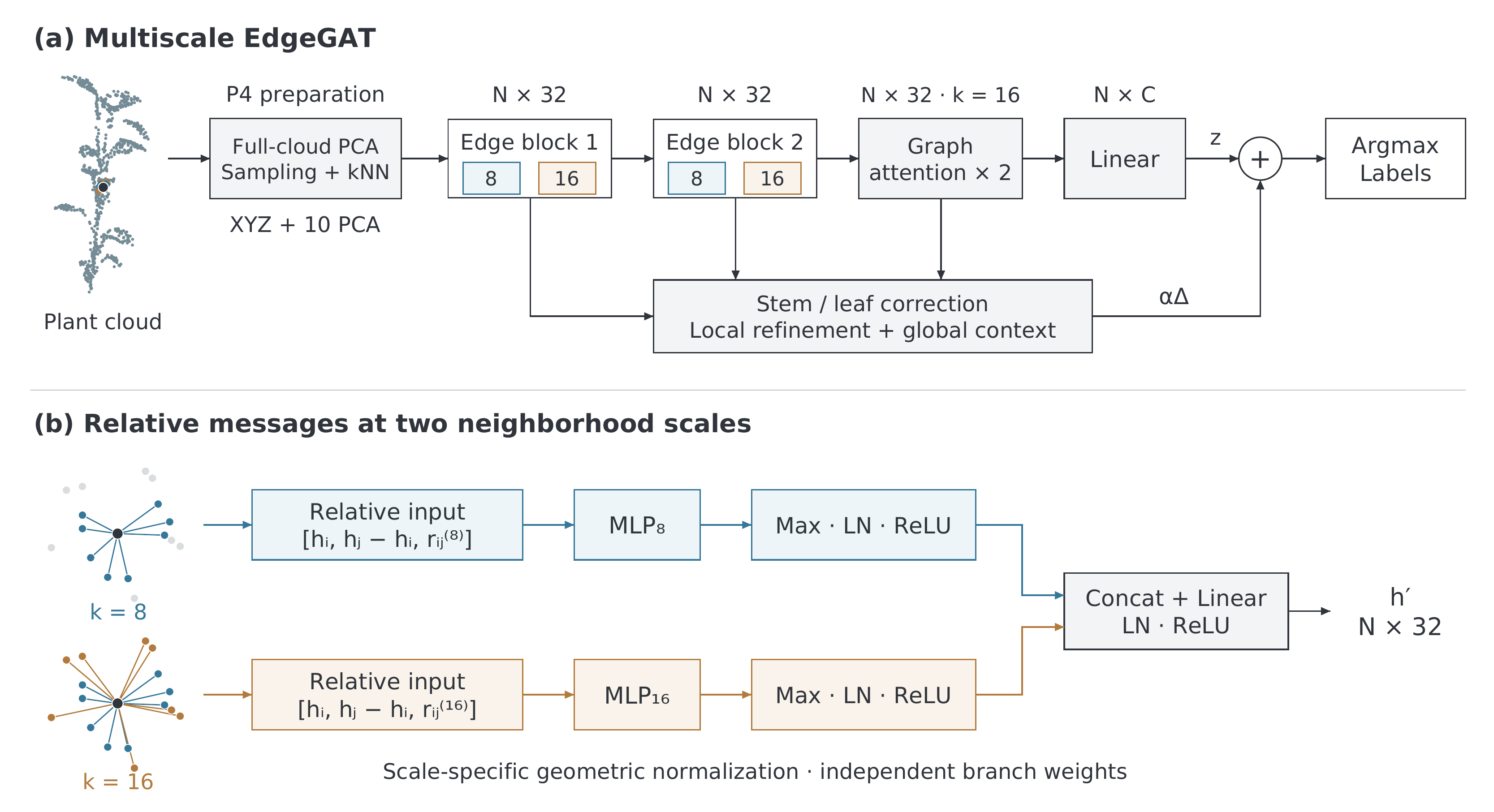}}
\caption{Multiscale EdgeGAT. (a) Full-cloud PCA descriptors, sampling and coordinate neighborhoods supply two multiscale edge blocks and two graph attention layers. P4 optimizes preparation without changing the trained network. A separate branch combines local refinement and global context to correct only stem and leaf logits; argmax assigns the final labels. (b) Each edge block computes relative inputs separately over nested neighborhoods of 8 and 16 points. The geometric term $r_{ij}^{(k)}$ uses a scale specific local normalization. Independently learned messages are aggregated by neighbor maxima and fused into 32 features per point. Blue marks the nearest eight neighbors; ochre marks the next eight, also included at $k=16$. The plant and neighborhoods show real geometry, not predicted labels. LN denotes layer normalization. The editable vector schematic was prepared with OpenAI Codex assistance.}
\label{fig:arch}\end{figure}

The model processes and labels 1,024 sampled points, with XYZ coordinates and ten local PCA descriptors. It does not assign labels to all points of the original cloud. A hidden width of 32 and two attention heads keep the learned network compact. Two relative edge feature blocks use neighborhoods of sizes 8 and 16, followed by graph attention. Global mean and maximum summaries provide cloud context. A residual branch corrects stem and leaf logits; an auxiliary stem head contributes during training and is disabled during inference. Figure~\ref{fig:arch} separates geometric preparation from the frozen inference network.

The frozen ternary AO reference has 77,863 parameters, a refiner with neighborhoods 12 and 32, and training dropout 0.05. The historical binary model deployed on PlantNet and TomatoWUR has 77,830 parameters, refiner neighborhoods 8 and 32, and dropout 0.1. These are distinct recorded recipes. The figure shows their shared inference flow; class counts and refiner settings remain specific to each recipe. Both retain the same multiscale backbone principle. No late hierarchy of sampled anchors is part of the model evaluated with P4.

An illustrative edge input is
\begin{equation}
 m_{ij}=\phi_\theta([h_i,h_j-h_i,r_{ij}]),\qquad
 r_{ij}=\left[\frac{x_j-x_i}{s_i},\frac{\lVert x_j-x_i\rVert_2}{s_i},q_{ij}\right],
\end{equation}
where $s_i$ is the clamped median neighbor distance and $q_{ij}$ contains PCA based relations. The implementation retains its existing aggregation and attention. This expression describes the frozen message input, not a newly introduced EdgeConv operator.

\subsection{What PCA contributes to the representation}
For each raw point, the descriptor extractor obtains 16 neighbors, centers their coordinates and forms a covariance matrix. With descending eigenvalues $\lambda_1\geq\lambda_2\geq\lambda_3$ after numerical clamping, the ten channels comprise the three components of the smallest eigenvector and seven scalar quantities: curvature $\lambda_3/\sum_j\lambda_j$, linearity, planarity, sphericity, omnivariance, anisotropy and eigenentropy. The scalar definitions and numerical regularization follow the original code.

Descriptor extraction covers the entire retained raw cloud before sampling. Feature columns undergo the original full cloud mean and standard deviation normalization, and sampled features enter the model's existing robust normalization. P4 preserves both stages. It does not substitute physical normals for standardized features, reduce the PCA support, approximate eigenvectors or move PCA after sampling. Those would change the representation and require a different experiment.

PCA makes local shape explicit, but its usefulness must be established through controlled accuracy comparisons. The E1 variant removes PCA information and retains a single neighborhood scale of 16 in both the edge blocks and refiner. It has 50,535 parameters with zero filled descriptor channels; a native XYZ inference representation removes redundant weight columns and has 49,383 parameters. E1 therefore changes geometry information and scale structure together. To separate these factors within this model family, the completed ablation adds an XYZ multiscale variant (E2) and a PCA single scale variant (E3). E2 retains the full model's 77,863 parameters; E3 retains E1's 50,535. The scale factor changes the backbone and refiner together, including their capacity, and is not a parameter matched comparison.

\subsection{Prediction preserving preparation}
The optimized implementation has three parts. First, a compiled C routine performs farthest point sampling, using the same candidate cap of 20,000, random candidate subset, first point, distance arithmetic and tie rule as the reference. Compiler options disable fast math and floating point contraction. The routine reduces interpreter and temporary array overhead without changing which 1,024 points are selected.

Second, PCA is evaluated in blocks of 4,096 points using four worker threads. Each block retains the original covariance and eigendecomposition expressions. Blocks are placed back into their original order before full cloud standardization. Linear algebra libraries use one thread so that parallelism is not multiplied inadvertently. Third, the external coordinate neighbor query uses four workers. The graph remains the original Euclidean graph with the same removal of the first returned neighbor; there is no change in neighborhood sizes or learned connectivity.

The optimized pipeline can be expressed as
\begin{equation}
 X\longrightarrow \{D(X),I(X;s)\}\longrightarrow
 [\widehat X_I,D(X)_I],\ G(\widehat X_I)\longrightarrow f_\theta\longrightarrow\widehat y,
\end{equation}
where $D$ is full cloud PCA extraction, $I$ is sampling with frozen seed $s$, $G$ is graph preparation and $\theta$ is a fixed trained checkpoint. P4 changes how $D$, $I$ and $G$ are computed. It does not change their intended values or $f_\theta$.

PointNeXt uses XYZ and constructs its own neighbors inside the network. It does not receive Multiscale's PCA or external graph optimization. Both models nevertheless require external reduction of dense clouds in this evaluation. We therefore retain a secondary PointNeXt measurement with the same compiled sampling routine. This identifies the shared sampling benefit and prevents attributing the entire acceleration to PCA or comparing optimized sampling only against an unnecessarily slow baseline.

\section{Experimental design}
\subsection{Collections and evidence levels}
The collection identified locally as AO derives from the field maize terrestrial LiDAR study of \citet{ao2022}. We retain the AO identifier to preserve correspondence with the archived configurations and results. The 304 processed clouds and seven filename groups in our local protocol are not counts of independent field samples from that publication.

Table~\ref{tab:data} summarizes seven local collection identifiers. All models are trained and evaluated within each collection; no transfer experiment is included. Binary labels are stem zero and leaf one. PlantNet merges its species and organ labels into this order. TomatoWUR merges main and side stems, retains leaves and excludes support poles and ignored points using the supplied annotations before sampling. This defines an annotation-prefiltered organ classification input, not autonomous plant/support separation. The evaluation does not measure how that preisolated population would be obtained from an unlabelled scene. The binary tasks cannot be compared directly with original dataset leaderboards.

\begin{table}[tbp]\centering\small
\caption{Local collection structure. Counts refer to files and protocol groups, not necessarily independent plants. Ternary labels are stem, soil and leaf; binary labels are stem and leaf.}\label{tab:data}
\begin{tabular}{lrrll}\toprule
Collection & Clouds & Groups & Classes & Group basis\\\midrule
AO & 304 & 7 & Ternary & Filename prefixes\\
Pheno4D subset & 120 & 40 & Ternary & Local clusters\\
ZeaMays & 2,080 & 520 & Ternary & Source identifiers\\
Single Maize & 428 & 5 & Binary & Cultivar holdouts\\
PLANesT\textendash3D & 34 & 31 & Binary & Reconstruction groups\\
PlantNet & 546 & 35 & Binary & Species and replicates\\
TomatoWUR & 44 & 44 & Binary & Pot identifiers\\\bottomrule
\end{tabular}
\end{table}

The evidence is separated into three levels. The nested AO study uses seven outer groups, a different group for inner checkpoint selection, and seeds 42, 52 and 62. Historical comparisons across all collections use grouped validation folds that also selected the training epoch. The P4 study is a paired inference comparison using existing seed 42 checkpoints on PlantNet and TomatoWUR, with one desktop cycle and three repeated ARM cycles. It evaluates implementation integrity and runtime; it neither retrains models nor creates new independent accuracy evidence. AO and Pheno4D have distinct published source studies \citep{ao2022,pheno4d}; the mapping of processed local files to source individuals and captures has not been fully documented. The public mapping of local ZeaMays files also remains incomplete. Seven identifiers must not be interpreted as seven certified independent cohorts. The source-to-individual-to-capture mapping needed to verify biological independence remains incomplete, and nested checkpoint selection does not resolve that provenance gap. AO was also used during development, so it is not a previously untouched confirmatory dataset.

\subsection{Completed nested factorial ablation}
The AO ablation crosses two factors: PCA information present or replaced by ten constant zero channels, and a single scale configuration or the frozen multiscale configuration. E1 is single scale XYZ; E2 is multiscale XYZ; E3 is single scale with PCA; Multiscale is the full configuration. All variants retain 1,024 points and the original losses. The scale intervention changes backbone and refinement neighborhoods together. It therefore estimates a configuration effect, not the isolated benefit of a second branch at fixed parameter count.

The study comprises 84 results: four variants, seven outer groups and seeds 42, 52 and 62. The 42 E1 and Multiscale controls are reused after artifact verification; 42 E2 and E3 jobs complete the same nested protocol. Inner validation selects the checkpoint before outer evaluation. Independent first step and resume checks passed in the archived CPU harness, and the pairing audit verifies sample order, XYZ inputs and labels over the compared epochs. PCA content is deliberately different between feature arms. Differences are computed within each seed and outer group, then seeds are averaged within groups. Descriptive 95\% intervals use 10,000 bootstrap resamples of the seven groups with seed 20261009. They are not adjusted for multiple contrasts and do not establish population wide superiority. The frozen design is provided as \texttt{factorial\_protocol.json}; \texttt{data\_ao\_protocol.json} belongs to the earlier AO multiseed phase and is not the factorial manifest.

\subsection{Broad benchmark and focused comparisons}
The first comparison uses the complete common seed 42 cohort, so every baseline in a given table contributes the same number of folds. On AO, Pheno4D and ZeaMays, it includes Multiscale, its late cloud context variant, PointNet++, DGCNN, Point Transformer, RandLA-Net, PSegNet with XYZ, PSegNet with PCA features and PointNeXt. AO contributes seven folds per recipe and the other collections five. PSegNet with PCA is a feature control, not a distinct architecture family; the context variant belongs to our model rather than the external baselines.

On Single Maize and PLANesT-3D, the recorded binary benchmark contains Multiscale and six external recipes: PointNet++, DGCNN, Point Transformer, RandLA-Net, PSegNet XYZ and PointNeXt. Each contributes five folds at seed 42. The plant extension has different coverage: Multiscale, PointNeXt, Point Transformer V3 and PointCT adapters, each with five folds and seeds 42, 52 and 62 on PlantNet and TomatoWUR. Missing combinations are not inferred or filled with zero. Point Transformer V3 uses an official backbone with local segmentation and compatibility settings, without pretrained weights. PointCT changes the original weak supervision setting to a local fully supervised task \citep{pointct}. Neither adapter is an official reproduction.

The broader common seed tables precede the focused three seed Multiscale versus PointNeXt summary. They expose the basis for selecting PointNeXt without mixing one seed secondary baselines with three seed primary means in the same ranking. The nested AO study then tests accuracy and geometric information under separate inner and outer groups, including E1, DGCNN and RandLA-Net in its seed 42 comparison. The inference studies address a further question: whether compactness translates into lower network and preparation cost. This ordering explains the evidence chain; comparator selection followed exploratory observations and was not a preregistered selection rule. All rankings concern the frozen local recipes and their recorded feature modes.

\subsection{Training context and metrics}
In nested AO, five groups train the model, the next cyclic group supplies inner validation and one group is reserved for outer evaluation. Training uses FP32, batch size two, AdamW with learning rate $10^{-3}$ and weight decay $10^{-5}$, a 50 epoch cosine schedule, gradient clipping at one and early stopping patience 15. Rotation about the vertical axis, scaling from 0.9 to 1.1 and Gaussian jitter with standard deviation 0.005 form the augmentation recipe. Shared explicit generators pair sample order and augmentations across models for common epochs.

All four factorial variants use cross entropy, 0.5 Lov\'asz loss \citep{lovasz}, 0.1 auxiliary stem binary cross entropy and 0.2 stem Tversky loss. The baselines use cross entropy plus 0.5 Lov\'asz loss. These frozen recipes differ in objective as well as architecture. Binary historical training settings are preserved in their own manifests; the AO recipe is not retroactively assigned to them. P4 has no training hyperparameters and performs no checkpoint selection.

Class intersection over union is $TP/(TP+FP+FN)$. Confusion counts cover only the 1,024 sampled labels per cloud and are pooled within each fold before class scores and mIoU are computed. Accuracy means weight folds equally. The earlier three-recipe AO paired contrasts in Table~\ref{tab:intervals}, archived in \texttt{data\_ao\_intervals.csv}, use 100,000 group bootstrap resamples with seed 20260925 after averaging three seeds within each of the seven groups. The separate four-variant factorial contrasts accompanying Table~\ref{tab:factorial}, archived in \texttt{ablation\_nested\_cluster\_bootstrap.csv}, use the 10,000 resamples specified above. These are distinct analyses, not two descriptions of one calculation. Small group counts, shared training sets and exploratory selection limit inference. Runtime instead weights observed clouds equally, which is a different aggregation and is stated explicitly. The AO boundary diagnostic marks stem or leaf points whose 16 coordinate neighbors include the other organ. Its cross confusion counts stem to leaf and leaf to stem predictions on that mask; errors into soil are excluded from the numerator. It is not an anatomical connectivity or distance metric.

\subsection{Paired desktop P4 protocol}
The host is an Intel Core i7 14700KF using four physical CPU cores, FP32 and batch size one. PyTorch uses four threads, while BLAS uses one. The software environment includes PyTorch 2.5.1, NumPy 1.26.4, SciPy 1.14.1 and PyG 2.6.1. The presence of a CUDA enabled host package does not mean these timing measurements use the GPU.

Twenty jobs cover two models, two datasets and five folds. Every cloud receives both the original and optimized preparations with the same checkpoint and seed. The sampling seed is $42+104729i$, where $i$ is the sorted validation index within the fold. Original and optimized order alternates across clouds and jobs. Two warmups per preparation precede measurement. The original instrumented path is checked against the frozen historical loader on the first and last cloud of each job.

The measured total begins with TXT reading and includes PCA, sampling, normalization, feature assembly, neighbor construction, tensor packing, inference and label selection for the 1,024 sampled points. Throughout the timing results, \emph{complete pipeline} denotes this bounded input-to-sampled-label path. It excludes dense label propagation and the acquisition of an annotation-free preisolated plant input. Hashing, sensor checks, numerical comparison and metric accumulation are outside the timer. No feature cache is used, but the operating system file cache is not cleared. Cold startup, model loading, sensing and reconstruction are not included. The desktop study contains one measured cycle and therefore describes this host observation, not a distribution over independent timing processes.

\subsection{Integrity and physical board evidence}
Each comparison checks hashes including dtype, shape and bytes for sample indices, coordinates, labels, features, neighbors and the complete PCA array. Predictions must match exactly, logits must meet absolute and relative tolerance $10^{-4}$, and bitwise equality is recorded separately. A failure stops the campaign without loosening tolerances. Actual desktop and ARM equality within each platform was stronger than the acceptance rule.

A previously completed physical Raspberry Pi 5 campaign evaluates the same 590 clouds and 20 checkpoints using the original preparation. The board has 8 GB RAM, active fan cooling, the user reported Pi power supply and an ARM64 CPU environment. Four threads and FP32 are used. Its timing scope is reported separately because its preparation plus network sum omits packing and postprocessing. A separate earlier board workload compared P4 on E1 and the ternary Multiscale model. These completed measurements are included with their own scope.

The completed paired P4 ARM study evaluates PlantNet and TomatoWUR on Raspberry Pi 5 over three repeated inference cycles. Each cycle comprises 20 model--dataset--fold jobs, for 60 jobs overall, using the same 20 seed 42 checkpoints and 590 clouds. Each job runs in a separate worker process; the three cycles do not constitute independent training or biological replication. PointNeXt receives the compatible FPS optimization. The timer includes TXT loading, preparation, packing, forward inference and argmax, and excludes warmup, verification, hashing and sensors. It retains the desktop study's sampled-label boundary, with no feature cache and no page-cache clearing. The older preparation-plus-network sums have a different scope and are retained separately.

The principal ARM latency is the mean across clouds of each cloud's median over three cycles. Its p50 and p95 describe the distribution of those per-cloud medians, not percentiles of independent process runs. Separately, means across clouds are calculated within each cycle; their mean, standard deviation and range describe variation across the three cycles. This second aggregation is not the center or uncertainty interval of the principal mean-of-medians. Stage medians are also summarized individually and need not sum to the separately measured total. Repeated cycles add timing observations but no new plants, folds, training seeds or independent accuracy evidence.

\section{Results}
\subsection{Broad comparison with segmentation baselines}
Table~\ref{tab:baselines3} presents all nine recorded ternary configurations at the common seed 42. PointNeXt leads the external baselines on AO (88.96\% mIoU), Pheno4D (80.11\%) and ZeaMays (90.51\%). The next external recipe depends on the collection: PSegNet XYZ on AO, and DGCNN on Pheno4D and ZeaMays. Thus the later focus on PointNeXt follows a comparison with several plausible alternatives. Multiscale is competitive on AO but falls below DGCNN and PointNeXt on the other two collections. Its late context variant reaches 89.36\% on AO, exceeding PointNeXt there, but does not lead on Pheno4D or ZeaMays.

\begin{table}[tbp]\centering\small
\caption{Broad ternary benchmark: mIoU in percent at seed 42, using all seven AO folds and all five folds on Pheno4D and ZeaMays. Checkpoints are validation selected. External recipes appear above the rule and our model configurations below it. Bold marks the highest mean in each column. PSegNet with PCA is a feature control.}\label{tab:baselines3}
\begin{tabular}{lrrr}
\toprule
Model & AO & Pheno4D & ZeaMays \\
\midrule
PointNet++ & 86.90 & 75.88 & 84.94 \\
DGCNN & 87.06 & 78.15 & 89.60 \\
Point Transformer & 87.58 & 75.49 & 88.93 \\
RandLA-Net & 83.77 & 74.27 & 84.30 \\
PSegNet XYZ & 88.10 & 76.71 & 88.20 \\
PSegNet + PCA & 87.04 & 70.99 & 87.56 \\
PointNeXt & 88.96 & \textbf{80.11} & \textbf{90.51} \\
\midrule
Multiscale & 88.78 & 75.82 & 88.66 \\
Multiscale + late context & \textbf{89.36} & 75.77 & 88.61 \\
\bottomrule
\end{tabular}
\end{table}

The binary benchmark in Table~\ref{tab:baselines2} includes the six external families as well as Multiscale. PointNeXt again has the highest recorded mean, reaching 95.34\% on Single Maize and 83.01\% on PLANesT-3D. Multiscale ranks second on both, at 95.03\% and 81.18\%, respectively. DGCNN and PSegNet are the next two recipes on these datasets. The lower results of the remaining recipes are retained rather than excluded from the comparison.

\begin{table}[tbp]\centering\small
\caption{Broad binary benchmark: mIoU in percent at seed 42, with all five folds per model and collection. These are validation selected checkpoints, distinct from the nested AO evaluation. Bold marks the highest mean in each column.}\label{tab:baselines2}
\begin{tabular}{lrr}
\toprule
Model & Single Maize & PLANesT 3D \\
\midrule
PointNet++ & 88.35 & 69.85 \\
DGCNN & 94.48 & 77.08 \\
Point Transformer & 90.03 & 72.67 \\
RandLA-Net & 86.27 & 66.82 \\
PSegNet XYZ & 93.61 & 76.78 \\
PointNeXt & \textbf{95.34} & \textbf{83.01} \\
\midrule
Multiscale & 95.03 & 81.18 \\
\bottomrule
\end{tabular}
\end{table}

The plant extension retains all four evaluated recipes (Table~\ref{tab:expansion}). PSegNet was not evaluated in this extension cohort, despite its relevance to these species; GCASSN, GSAN, ELGCot3D, PlantEFRSegnet and LitePT are literature comparators only. Their absence limits a current state-of-the-art claim. PointNeXt leads the external adapters on both collections. On PlantNet, Multiscale reaches 90.39\% against 90.35\% for PointNeXt; the unrounded difference is only 0.033 percentage points and does not establish superiority. Both outperform the recorded Point Transformer V3 and PointCT adapters. On TomatoWUR, PointNeXt reaches 80.53\%, Point Transformer V3 77.00\% and Multiscale 74.62\%. The stem IoU difference between Multiscale and PointNeXt is substantial. This limitation is as relevant as the favorable PlantNet result.

\begin{table}[tbp]\centering\small
\caption{Plant extension: all four evaluated recipes, each with five folds and three seeds (42, 52 and 62), giving 15 fold results per row. Scores are percentages from validation selected checkpoints. Bold marks the highest mIoU per collection. The adapters use local training and compatibility settings.}\label{tab:expansion}
\begin{tabular}{llrrr}
\toprule
Collection & Model & mIoU & Stem IoU & Leaf IoU \\
\midrule
PlantNet & PointNeXt & 90.35 & 85.07 & 95.64 \\
PlantNet & PTv3 adapter & 88.28 & 81.89 & 94.67 \\
PlantNet & PointCT adapter & 63.01 & 43.61 & 82.41 \\
PlantNet & Multiscale & \textbf{90.39} & 85.15 & 95.63 \\
\midrule
TomatoWUR & PointNeXt & \textbf{80.53} & 67.88 & 93.18 \\
TomatoWUR & PTv3 adapter & 77.00 & 61.85 & 92.15 \\
TomatoWUR & PointCT adapter & 60.00 & 32.48 & 87.52 \\
TomatoWUR & Multiscale & 74.62 & 58.00 & 91.23 \\
\bottomrule
\end{tabular}
\end{table}

\begin{figure}[tbp]\centering\includegraphics[width=\linewidth]{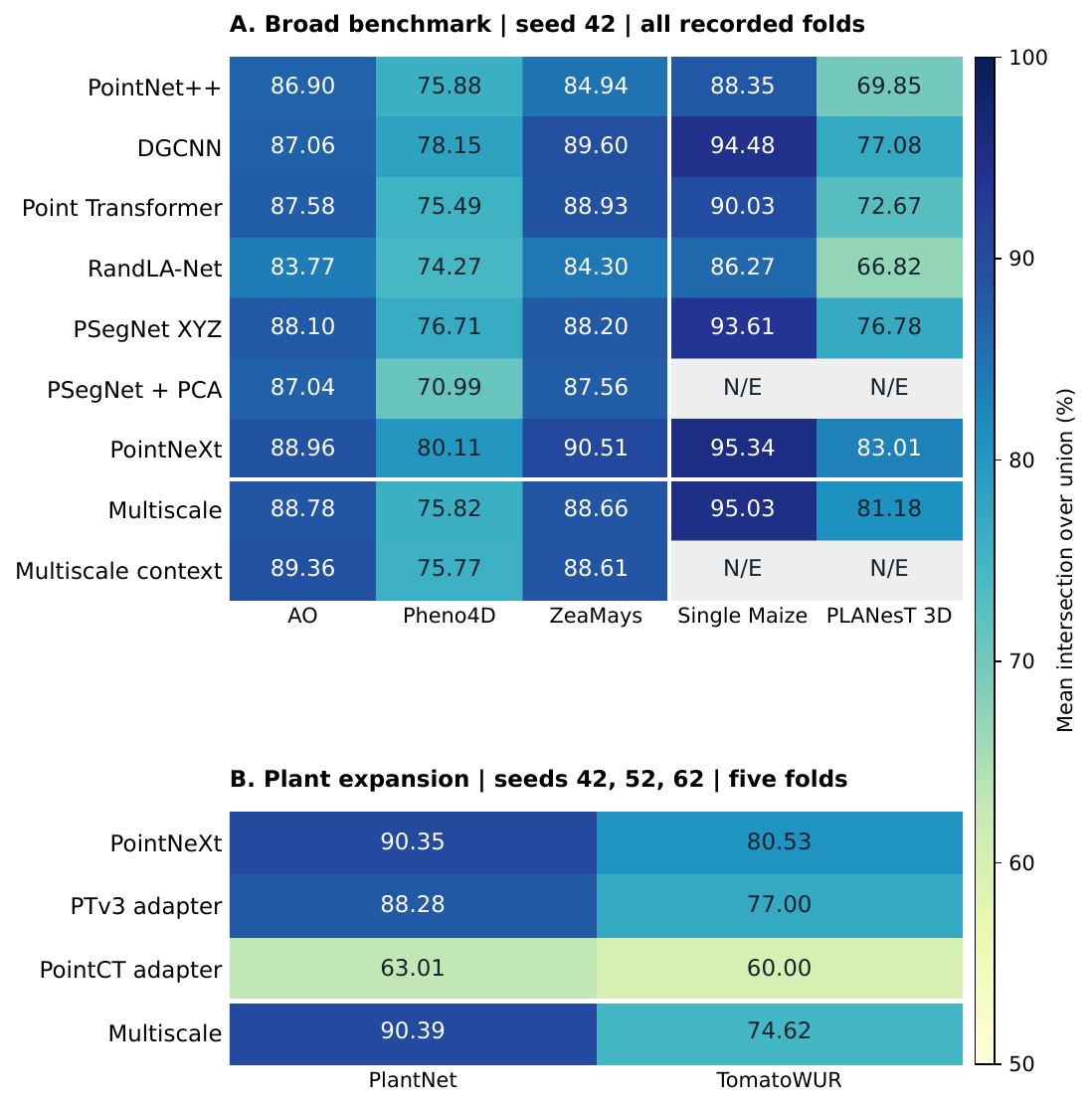}
\caption{Overview of the complete available baseline cohorts. Panel A uses seed 42 with seven AO folds and five folds elsewhere; its first three collections are ternary and its last two binary. Panel B shows the separate binary plant extension with five folds and three seeds. Cells contain mIoU in percent; N/E means not evaluated in that cohort. Colors use one fixed scale, without pooling datasets, tasks or seed cohorts. Generated from archived CSV tables with an AI assisted plotting script.}\label{fig:accuracy}\end{figure}

\subsection{Why PointNeXt becomes the focused reference}
Figure~\ref{fig:accuracy} makes the basis for the focused comparison visible. Among the external recipes actually tested on each collection, PointNeXt has the highest mean in every displayed cohort. This supports choosing it as the subsequent accuracy reference, while retaining DGCNN and RandLA-Net in the wider nested AO check. The observation does not make PointNeXt the best model in every comparison: our late context variant leads on AO at seed 42, and Multiscale has a slightly higher mean on PlantNet.

Table~\ref{tab:multi} then reports the available three seed comparison between Multiscale and PointNeXt across all seven collection identifiers. PointNeXt has higher mIoU on six; PlantNet is the small exception. Multiscale is close on AO and Single Maize, whereas the gaps on Pheno4D and TomatoWUR are larger. This focused table extends the broad accuracy context and leads to the engineering question: what storage and runtime advantages does a much smaller model provide at its observed accuracy? It does not replace the other baseline results.

\begin{table}[tbp]\centering\small
\caption{Focused three seed comparison after the broad benchmark. Historical validation selected mIoU in percent, using all folds and seeds 42, 52 and 62. Runs are results per model. Differences use unrounded means. The PlantNet and TomatoWUR rows reuse the preceding extension cohort, rather than adding independent evidence.}\label{tab:multi}\resizebox{\linewidth}{!}{\begin{tabular}{lrrrrr}
\toprule
Collection & Classes & Runs & Multiscale & PointNeXt & $\Delta$ (pp) \\
\midrule
AO & 3 & 21 & 88.29 & 88.71 & -0.41 \\
Pheno4D & 3 & 15 & 75.84 & 80.00 & -4.17 \\
ZeaMays & 3 & 15 & 88.67 & 90.47 & -1.80 \\
PLANesT-3D & 2 & 15 & 80.63 & 82.69 & -2.06 \\
Single Maize & 2 & 15 & 94.84 & 95.37 & -0.53 \\
PlantNet & 2 & 15 & 90.39 & 90.35 & +0.03 \\
TomatoWUR & 2 & 15 & 74.62 & 80.53 & -5.92 \\
\bottomrule
\end{tabular}
}\end{table}

The late cloud context adapter and geometric logit residual remain separate from P4. The former adds a hierarchical context correction; the latter uses PCA compatibility on a local graph to adjust logits. Their effects vary across datasets and do not justify silently replacing the frozen model. Table~\ref{tab:context} retains this architectural context in the appendix.

\subsection{Nested accuracy and the compact variants}
The nested AO study follows with separate groups for checkpoint selection and outer evaluation. Its broader seed 42 comparison (Table~\ref{tab:seed42}) includes E1, Multiscale, DGCNN, RandLA-Net and PointNeXt. Multiscale, DGCNN and PointNeXt have similar mIoU in this cohort, while PointNeXt has the highest stem IoU. These results show why the broad historical ranking should not be treated as a universal ordering across protocols.

\begin{table}[tbp]\centering\small
\caption{Broader nested AO comparison at seed 42 only. Seven outer groups contribute to each row; inner groups select checkpoints. Overlapping rows in the following three seed table are not additional independent evidence.}\label{tab:seed42}\begin{tabular}{lrrrr}
\toprule
Model & mIoU & Stem IoU & Soil IoU & Leaf IoU \\
\midrule
E1 & 86.42 & 66.64 & 98.97 & 93.66 \\
Multiscale & 86.74 & 66.84 & 99.36 & 94.04 \\
DGCNN & 86.48 & 66.39 & 99.39 & 93.66 \\
RandLA-Net & 80.90 & 53.98 & 97.90 & 90.80 \\
PointNeXt & 86.73 & 67.22 & 99.40 & 93.56 \\
\bottomrule
\end{tabular}
\end{table}

Table~\ref{tab:ao} gives the three seed nested AO results. Multiscale reaches 86.36\% mIoU and 66.55\% stem IoU. E1 reaches 85.71\% and 65.18\%, respectively. The unrounded differences favor Multiscale by 0.64 and 1.36 percentage points. PointNeXt reaches 86.76\% mIoU and 67.06\% stem IoU. Table~\ref{tab:intervals} gives the paired intervals. Thus the smaller models provide useful operating points, but the paired intervals do not establish equivalence to each other or to PointNeXt.

\begin{table}[tbp]\centering\small
\caption{Nested AO accuracy in percent, averaging seven outer groups and three paired seeds. Checkpoints are selected on separate inner groups.}\label{tab:ao}\begin{tabular}{lrrrr}
\toprule
Model & mIoU & Stem IoU & Soil IoU & Leaf IoU \\
\midrule
E1 & 85.71 & 65.18 & 98.75 & 93.20 \\
Multiscale & 86.36 & 66.55 & 98.84 & 93.68 \\
PointNeXt & 86.76 & 67.06 & 99.54 & 93.69 \\
\bottomrule
\end{tabular}
\end{table}
\begin{table}[tbp]\centering\footnotesize
\caption{Descriptive paired AO differences in percentage points, with 95\% group bootstrap intervals. Seeds are averaged before resampling. Positive boundary error differences are unfavorable.}\label{tab:intervals}\resizebox{\linewidth}{!}{\begin{tabular}{llrr}
\toprule
Comparison & Measure & Difference (pp) & 95\% interval \\
\midrule
E1 $-$ Multiscale & mIoU & -0.64 & [-1.68, 0.39] \\
E1 $-$ Multiscale & Stem IoU & -1.36 & [-3.97, 1.21] \\
E1 $-$ Multiscale & Boundary error & 0.51 & [-1.14, 2.41] \\
Multiscale $-$ PointNeXt & mIoU & -0.41 & [-2.65, 2.46] \\
Multiscale $-$ PointNeXt & Stem IoU & -0.51 & [-5.55, 5.44] \\
Multiscale $-$ PointNeXt & Boundary error & 4.08 & [1.46, 7.12] \\
\bottomrule
\end{tabular}
}\end{table}

\subsection{Separating geometric information and scale configuration}
Table~\ref{tab:factorial} completes the nested factorial comparison. Multiscale has the highest mean mIoU (86.36\%) and stem IoU (66.55\%) among the four variants. Relative to the multiscale XYZ control, retaining PCA increases mIoU by 0.72 percentage points, with a descriptive interval of $[-0.16,1.63]$. Under the single scale configuration, the PCA difference is $-0.10$ points, with an interval of $[-1.66,1.40]$. These results do not establish an unconditional accuracy advantage of PCA.

\begin{table}[tbp]\centering\small
\caption{Completed nested AO factorial ablation. Values are percentages averaged over seven outer groups and three seeds. Lower boundary error is better. E1 and Multiscale reuse the controls reported above; the rows are not additional independent evidence. XYZ variants retain ten constant zero descriptor channels during training.}\label{tab:factorial}
\resizebox{\linewidth}{!}{\begin{tabular}{llrrrrr}\toprule
Model & Configuration & Params. & mIoU & Stem IoU & Leaf IoU & Boundary error\\\midrule
E1 & Single scale, XYZ & 50,535 & 85.71 & 65.18 & 93.20 & 27.91\\
E2 & Multiscale, XYZ & 77,863 & 85.64 & 65.59 & 92.91 & 26.10\\
E3 & Single scale, PCA & 50,535 & 85.62 & 64.89 & 93.17 & 29.17\\
Multiscale & Multiscale, PCA & 77,863 & 86.36 & 66.55 & 93.68 & 27.40\\
\bottomrule\end{tabular}
}
\end{table}

With PCA present, the multiscale configuration increases stem IoU by 1.66 points, with an interval of $[0.03,4.01]$. This is one of several exploratory contrasts without multiplicity correction and is not a confirmatory component test. However, E2 has the lowest stem and leaf boundary error, 26.10\%, compared with 27.40\% for Multiscale. Adding PCA to the multiscale configuration raises this error by 1.30 points, with an interval of $[0.16,2.44]$. The full model's higher organ overlap therefore coexists with a less favorable local boundary result. Neither metric should substitute for the other. The mIoU interaction estimate is 0.81 points, with an interval of $[-0.18,2.12]$, so the data do not establish a general synergy effect. Complete contrasts are retained in the accompanying tables.

E1 remains a smaller operating point, with 50,535 training parameters and an equivalent 49,383 parameter native XYZ inference form. The completed ablation supports reporting the accuracy and cost tradeoff; it does not demonstrate that every component is necessary. P4 addresses the separate implementation question of retaining the full representation while reducing its cost.

\subsection{P4 preserves predictions}
All 20 desktop jobs pass. The 546 PlantNet clouds and 44 TomatoWUR clouds, each evaluated with two models, produce 1,180 paired comparisons. Every recorded input and full PCA hash is exact, and logits and predictions are bitwise identical between original and optimized preparation. Confusion matrices also match the archived CPU reference on each fold. The result establishes preservation for these checkpoints, inputs and this platform; it is not a proof for arbitrary numerical libraries or future hardware.

\begin{table}[tbp]\centering\small
\caption{Seed 42 CPU accuracy under original and optimized preparation. Both arms have identical values. Means and standard deviations use five folds; organ columns are mean IoU.}\label{tab:p4accuracy}
\begin{tabular}{llrrr}
\toprule
Dataset & Model & mIoU $\pm$ SD & Stem IoU & Leaf IoU \\
\midrule
PlantNet & Multiscale & 90.34 $\pm$ 2.61 & 85.10 & 95.59 \\
PlantNet & PointNeXt & 90.30 $\pm$ 2.31 & 84.96 & 95.65 \\
TomatoWUR & Multiscale & 74.33 $\pm$ 1.80 & 57.55 & 91.11 \\
TomatoWUR & PointNeXt & 80.50 $\pm$ 1.59 & 67.87 & 93.13 \\
\bottomrule
\end{tabular}
\end{table}

Table~\ref{tab:p4accuracy} reports the preserved accuracy. The PlantNet difference between Multiscale and PointNeXt is only 0.039 percentage points in this cohort. TomatoWUR favors PointNeXt by 6.167 points of mIoU and about 10.31 points of stem IoU. These values differ slightly from the historical three seed averages because the present table fixes seed 42 and uses CPU inference. The two tables must not be described as improvement or degradation over time.

\subsection{Where the preparation becomes faster}
\begin{figure}[tbp]\centering\includegraphics[width=\linewidth]{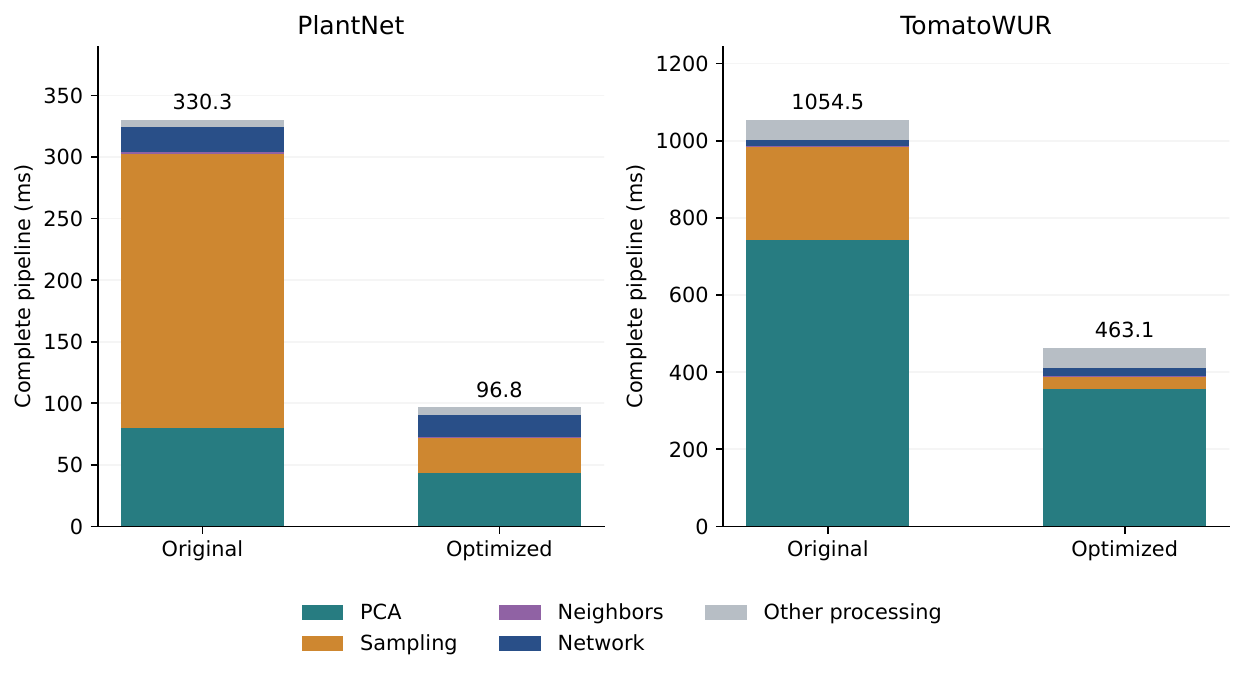}
\caption{Measured Multiscale desktop cost by stage, using separate scales for the two datasets. Other processing includes reading, normalization, feature assembly, packing, label selection and timed overhead. PCA retains the complete raw cloud support. Generated from recorded timings with an AI assisted plotting script.}\label{fig:stages}\end{figure}

\begin{figure}[tbp]\centering\includegraphics[width=\linewidth]{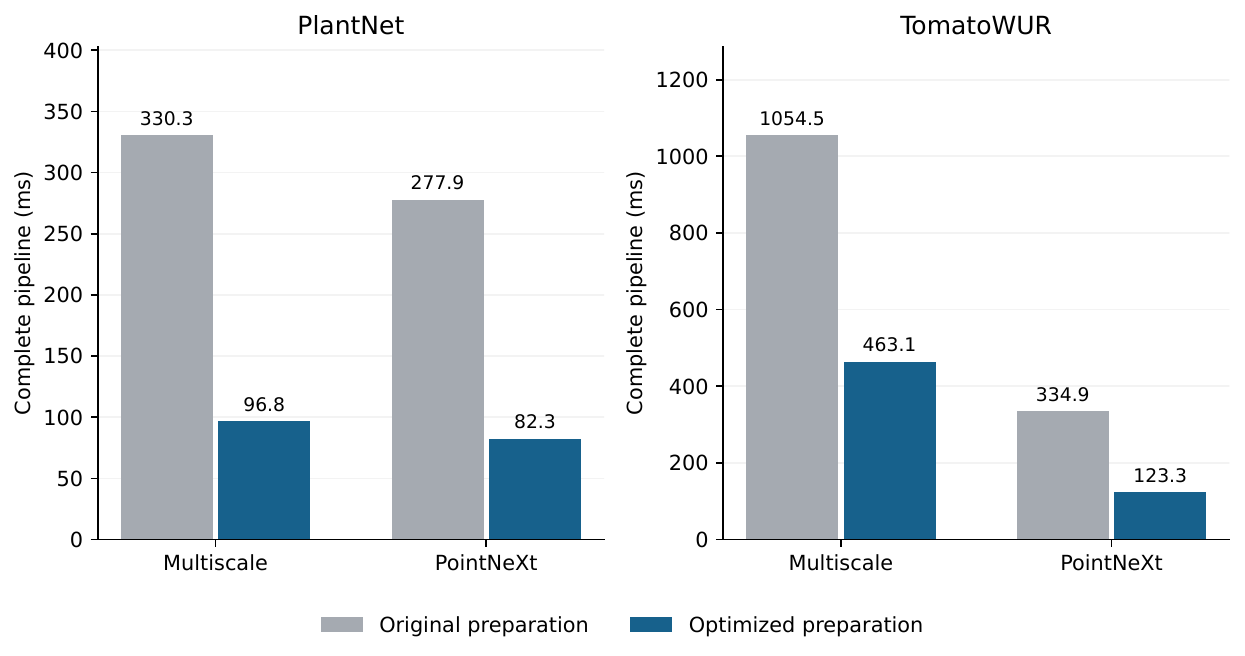}
\caption{Complete desktop pipeline means for both models, with original and optimized preparation. PointNeXt receives the compatible sampling optimization. Its absence of PCA is part of its actual recipe, not a missing timing component. Generated from recorded timings with an AI assisted plotting script.}\label{fig:fair}\end{figure}

Table~\ref{tab:p4time} and Fig.~\ref{fig:stages} show paired desktop runtime. Figure~\ref{fig:fair} also displays the compatible baseline optimization. Multiscale total latency falls by 70.69\% on PlantNet and 56.08\% on TomatoWUR, corresponding to factors of 3.41 and 2.28. The learned model is unchanged, so these are pipeline implementation gains rather than a faster neural architecture.

\begin{table}[tbp]\centering\small
\caption{Paired desktop pipeline latency, in milliseconds per cloud. PointNeXt optimization affects external sampling only; it has no external PCA or graph preparation in this protocol. One cycle is measured on four physical CPU cores.}\label{tab:p4time}\begin{tabular}{llrrr}
\toprule
Dataset & Model & Original ms & Optimized ms & Speedup \\
\midrule
PlantNet & Multiscale & 330.33 & 96.82 & 3.41$\times$ \\
PlantNet & PointNeXt & 277.93 & 82.34 & 3.38$\times$ \\
TomatoWUR & Multiscale & 1,054.52 & 463.10 & 2.28$\times$ \\
TomatoWUR & PointNeXt & 334.89 & 123.26 & 2.72$\times$ \\
\bottomrule
\end{tabular}
\end{table}

On PlantNet, PCA decreases from 79.88 to 43.00 ms, sampling from 222.53 to 28.42 ms and external neighbors from 1.83 to 1.16 ms. On TomatoWUR, the corresponding changes are 742.84 to 357.21 ms, 241.75 to 32.18 ms and 2.18 to 1.31 ms. Sampling supplies most of the absolute saving on PlantNet; PCA supplies the largest saving on TomatoWUR. The graph query is accelerated but represents a small fraction of the total at 1,024 retained points. It would be inaccurate to claim that graph construction dominates this workload.

After optimization, PCA still accounts for approximately 44.4\% of PlantNet total time and 77.1\% of TomatoWUR total time. P4 reduces its cost rather than eliminating it. Network stage times also vary between paired calls even though inputs and weights match; this is timing variation, not a model intervention. Only one desktop cycle is available, so the reported factors should not be given statistical confidence intervals based on treating clouds as independent hardware repetitions.

The fairness measurement matters. Compiled sampling reduces PointNeXt totals to 82.34 ms on PlantNet and 123.26 ms on TomatoWUR. On this desktop host, Multiscale with P4 is therefore not the fastest full pipeline when both models receive optimized sampling, despite its smaller and faster network. A comparison against PointNeXt's original 277.93 and 334.89 ms alone would hide this distinction. The defensible result is reduced geometric overhead with retained predictions and approximately 49.8 times fewer parameters than the binary local PointNeXt, not universal runtime superiority.

\subsection{Earlier Raspberry Pi reference evidence}
The original preparation campaign completed all 20 jobs on the physical board. Predictions match the CPU reference for both models on all 590 clouds. Multiscale's network profiles average 88.70 ms on PlantNet and 90.97 ms on TomatoWUR, compared with 262.72 and 260.75 ms for PointNeXt. However, preparation reverses that advantage at pipeline level (Table~\ref{tab:pi}). This directly motivates optimization instead of assuming that a compact network alone provides an efficient deployed system.

\begin{table}[tbp]\centering\small
\caption{Completed Raspberry Pi 5 reference campaign without P4. Network is the mean of five fold profile medians; preparation and preparation plus network are means over clouds. The sum omits packing and postprocessing and must not be directly subtracted from the new desktop total. RSS includes the audit runtime.}\label{tab:pi}
\begin{tabular}{llrrrr}\toprule
Dataset & Model & Network ms & Prep. ms & Sum ms & RSS MiB\\\midrule
PlantNet & Multiscale & 88.70 & 673.97 & 765.47 & 383.55\\
PlantNet & PointNeXt & 262.72 & 469.74 & 732.86 & 386.59\\
TomatoWUR & Multiscale & 90.97 & 2,498.75 & 2,590.63 & 426.13\\
TomatoWUR & PointNeXt & 260.75 & 611.83 & 873.14 & 378.88\\\bottomrule
\end{tabular}\end{table}

The completed reference run records 274 sensor readings, a maximum temperature of 68.1 degrees Celsius and no throttling flags. It validates physical operation during that campaign, not energy consumption. Numerical portability is also more qualified than label agreement. Multiscale's PCA features differ between CPU and ARM on all 590 clouds, although its logits meet the recorded tolerance. Six PointNeXt clouds fail strict logit tolerance across platforms, with maximum absolute difference 0.084638, while labels remain identical. These are retained failures of strict numerical equivalence; the responsible operation has not been isolated.

Earlier completed paired board measurements on a different mixed workload found E1 native preparation plus inference decreased from 393.11 to 100.29 ms and ternary Multiscale from 679.32 to 237.76 ms with P4. These observations support board feasibility of the optimization. They are not PlantNet or TomatoWUR measurements and their speedup factors must not be transferred to those datasets. The following paired study now provides direct measurements on these two collections with its own matched control and full timer.

\subsection{Completed paired P4 comparison on Raspberry Pi}
All 60 ARM jobs pass the frozen integrity gates. Across three cycles, 1,180 model--cloud pairs per cycle yield 3,540 implementation comparisons; these are repeated observations of the same 590 clouds. Inputs, logits and predictions are exact between original and P4 within ARM, with maximum absolute logit difference zero. Predictions are consistent across cycles, and both implementations retain identical confusion matrices. The five-fold seed 42 accuracy values match Table~\ref{tab:p4accuracy}: the PlantNet difference is approximately 0.039 percentage points and does not establish superiority. These repeated inference results neither improve accuracy nor provide an independent outer test.

\begin{figure}[tbp]\centering\includegraphics[width=\linewidth]{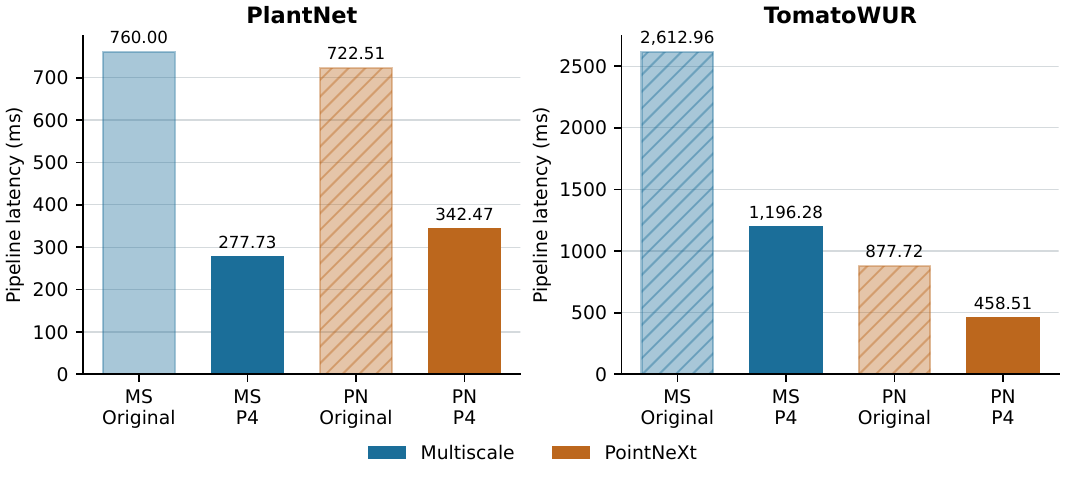}
\caption{Measured original and P4 pipeline latency on Raspberry Pi 5, with separate scales for each collection. Bars are means across clouds of per-cloud medians over three repeated cycles, not distributions over independent plants or training runs. PointNeXt receives optimized sampling. Generated from the verified ARM timing CSV with an AI assisted plotting script.}\label{fig:p4pi}\end{figure}

Table~\ref{tab:p4pi} reports the directly measured ARM comparison. Multiscale decreases from 760.00 to 277.73 ms on PlantNet and from 2,612.96 to 1,196.28 ms on TomatoWUR, reductions of 63.46\% and 54.22\%. PointNeXt decreases from 722.51 to 342.47 ms and from 877.72 to 458.51 ms, respectively, with the shared sampling optimization. Figure~\ref{fig:p4pi} shows these full pipeline means. No historical partial timer is substituted for the control arm.

\begin{table}[tbp]\centering\small
\caption{Paired Raspberry Pi 5 pipeline latency. Control and P4 are means across clouds of per-cloud medians over three cycles, in ms. Speedup is the ratio of those means. P4 p50 and p95 are percentiles across the per-cloud medians. Each row uses five folds and training seed 42.}\label{tab:p4pi}
\resizebox{\linewidth}{!}{\begin{tabular}{llrrrrrr}\toprule
Collection & Model & Control ms & P4 ms & Speedup & Reduction (\%) & P4 p50 ms & P4 p95 ms\\\midrule
PlantNet & Multiscale & 760.00 & 277.73 & 2.74$\times$ & 63.46 & 254.45 & 438.05\\
PlantNet & PointNeXt & 722.51 & 342.47 & 2.11$\times$ & 52.60 & 345.31 & 372.03\\
TomatoWUR & Multiscale & 2,612.96 & 1,196.28 & 2.18$\times$ & 54.22 & 1,179.17 & 1,592.76\\
TomatoWUR & PointNeXt & 877.72 & 458.51 & 1.91$\times$ & 47.76 & 457.18 & 503.76\\
\bottomrule\end{tabular}
}\end{table}

The mean latency ranking depends on the measured platform and workload. On PlantNet, Multiscale with P4 averages 277.73 ms on Pi versus PointNeXt's 342.47 ms; the corresponding desktop means are 96.82 and 82.34 ms, reversing the ordering. However, the PlantNet Pi p95 favors PointNeXt, 372.03 ms versus 438.05 ms for Multiscale. On TomatoWUR, PointNeXt remains faster on both platforms. One desktop cycle and three ARM cycles do not isolate hardware as a causal factor or establish a universal speed ranking. The observed advantage concerns a particular aggregation and workload.

Stage timings explain the remaining cost. Multiscale's ARM PCA stage decreases from 199.59 to 101.28 ms on PlantNet and from 1,896.86 to 897.19 ms on TomatoWUR. Its FPS stage decreases from 451.36 to 68.21 ms and from 497.09 to 77.78 ms, respectively. The full-cloud descriptor support is unchanged; the much larger PCA cost on TomatoWUR remains a deployment constraint. Network-stage fluctuations are timing observations with unchanged weights and inputs, not an architectural intervention.

\begin{table}[tbp]\centering\small
\caption{Descriptive variation across three ARM cycle means, in ms. For each cycle, latency is first averaged across clouds; the table then summarizes those three means. This aggregation differs from Table~\ref{tab:p4pi}, so its SD must not be attached to that table's principal mean-of-medians. Cycles repeat the same inference workload.}\label{tab:p4picycles}
\resizebox{\linewidth}{!}{\begin{tabular}{lllrrr}\toprule
Collection & Model & Preparation & Mean of cycle means & SD of cycle means & Range of cycle means\\\midrule
PlantNet & Multiscale & Original & 761.19 & 4.88 & 755.65--764.87\\
PlantNet & Multiscale & P4 & 278.94 & 0.32 & 278.68--279.29\\
PlantNet & PointNeXt & Original & 723.48 & 1.51 & 721.96--724.98\\
PlantNet & PointNeXt & P4 & 342.65 & 1.99 & 340.36--343.97\\
TomatoWUR & Multiscale & Original & 2,602.77 & 16.63 & 2,583.95--2,615.47\\
TomatoWUR & Multiscale & P4 & 1,195.89 & 4.37 & 1,190.98--1,199.37\\
TomatoWUR & PointNeXt & Original & 876.36 & 4.04 & 872.44--880.51\\
TomatoWUR & PointNeXt & P4 & 458.51 & 1.95 & 456.33--460.09\\
\bottomrule\end{tabular}
}\end{table}

Table~\ref{tab:p4picycles} retains the cycle-level timing variation without confidence intervals or independence claims. The completed P4 campaign records a maximum temperature of 66.4 degrees Celsius and only zero throttling flags. These observations establish operation under the recorded conditions; energy was not measured. Exact within-ARM equality does not erase the earlier cross-platform feature and logit discrepancies reported above.

\section{Discussion}
\subsection{From broad accuracy evidence to a deployment question}
The benchmark has two roles. First, it places Multiscale among hierarchical, graph, attention and plant oriented alternatives across distinct plant collections. Second, it identifies PointNeXt as a demanding external reference for the focused efficiency study. PointNeXt leads the tested external recipes in the displayed broad cohorts and exceeds Multiscale on six of the seven historical three seed comparisons. Accordingly, compactness is evaluated alongside an acknowledged accuracy tradeoff, rather than presented as evidence of better segmentation. The AO context variant and the near tie on PlantNet qualify that overall pattern.

Restricting the P4 comparison to Multiscale and PointNeXt does not imply that the other models were never tested or that they could not be deployed. The broad tables document the accuracy comparisons; the paired inference experiment measures an implementation change in our geometric pipeline and the shared sampling benefit in the strongest recorded external reference. No new latency ranking is assigned to baselines that were absent from that paired P4 workload. A comparison with PSegNet on PlantNet, a close plant graph method such as GCASSN, and a further efficient model under a shared protocol remains future work, not an experiment completed in this revision.

\subsection{A defensible role for geometric descriptors}
The study supports a practical distinction between representing local shape and paying unnecessary implementation overhead. Multiscale's established recipe retains explicit geometric descriptors and has broad historical evaluation. P4 computes the same representation more efficiently and preserves every prediction in the completed paired desktop and ARM cohorts within each platform. This is useful engineering evidence even though it is not a new PCA estimator or message passing theory.

The completed nested factorial comparison qualifies the role of geometry. PCA has a positive mean mIoU effect within the multiscale configuration, but its interval crosses zero, and the same information does not improve the single scale mean. The multiscale XYZ control also has lower boundary error than the full model. Thus the evidence supports a measured design tradeoff rather than a claim that PCA is necessary or uniformly beneficial. The preparation results answer the narrower question requested by an embedded implementation: an existing representation can be retained while its observed runtime is reduced. A future change to PCA support or approximation would require fresh accuracy evaluation and should not inherit the present equality claim.

\subsection{Compactness and practical deployment}
The binary Multiscale model has 77,830 parameters compared with 3,872,770 for the local PointNeXt. This reduces learned storage requirements, but the Raspberry process still occupies hundreds of MiB, and Multiscale has higher RSS than PointNeXt on TomatoWUR in the reference campaign. Network size, full pipeline latency and process memory are separate quantities.

Physical execution and a paired P4 benefit for both models are now demonstrated on these two collections. The resulting mean latency ranking is specific to the platform and workload, and PlantNet illustrates that a lower mean can coexist with a higher p95. An embedded agricultural product still needs more than a successful inference service. Sensor integration, dense reconstruction, variable point density, occlusion, wind and moving viewpoints remain outside the evaluated inputs. Power, battery endurance and payload mass have not been measured. Raspberry timings cannot establish Jetson performance, and cloud processing rates cannot be described as flight frame rates.

Multiscale's PlantNet performance is encouraging for plants beyond maize, while TomatoWUR identifies a substantial stem segmentation weakness. E1 remains a candidate with strong compactness evidence on AO, without equivalent trained evaluations on these two additional collections. The model should be selected against an explicit organ accuracy requirement and platform budget rather than a blanket claim that larger models cannot run on an embedded board.

\subsection{Validity and limitations}
The principal limitations are protocol heterogeneity, incomplete source mapping for local collections, and the use of local baseline recipes. The nested AO comparison differs from historical validation selected scores and from the inference only P4 study. Model selection and focused extensions followed exploratory results. The figures therefore show descriptive comparisons, not a preregistered proof of superiority. The scale intervention also changes backbone/refiner capacity, and external baselines use different auxiliary losses. A matched capacity and loss control would be needed to isolate a causal architectural benefit. Neither additional seeds nor the P4 checkpoint reuse removes these limitations.

The P4 dataset extension has one measured desktop cycle, three repeated ARM cycles, a fixed training seed and no new training. ARM cycle means provide descriptive timing variation, but three cycles do not characterize all operating conditions, and desktop variation across repeated cycles remains unmeasured. The paired ARM total uses the same bounded input-to-sampled-label scope as the desktop study; the older ARM reference sums and mixed workload retain their different scopes. Preserved predictions apply to the audited inputs within each platform. Historical cross-platform logit discrepancies remain unresolved, and neither an energy benefit nor autonomous scene preparation was measured.

\section{Conclusions}
The broad benchmark places Multiscale EdgeGAT against all evaluated external baseline recipes before the focused deployment study. PointNeXt provides the strongest external accuracy reference in the available cohorts, while the late context result on AO and the small Multiscale lead on PlantNet retain important exceptions. Multiscale offers a compact plant segmentation model whose engineering cost must include geometric preparation. The P4 implementation preserves full cloud PCA and the existing graph while reducing complete desktop pipeline latency by 70.69\% on PlantNet and 56.08\% on TomatoWUR. All 1,180 desktop comparisons and 3,540 repeated ARM comparisons retain exact logits and predictions within their respective platforms. The completed nested ablation adds controlled evidence about the design: the full configuration has the highest mean organ overlap, whereas multiscale XYZ yields lower boundary error. PCA effects depend on configuration and remain uncertain for mIoU. The contribution is consequently a compact evaluated design and lower implementation cost for its frozen representation, rather than universal component superiority.

The completed Raspberry Pi study directly establishes P4 reductions of 63.46\% on PlantNet and 54.22\% on TomatoWUR for Multiscale. With optimized sampling also available to PointNeXt, Multiscale has lower mean pipeline latency on PlantNet on Pi, whereas PointNeXt is faster on desktop, on TomatoWUR on both platforms, and at the PlantNet Pi p95. These are observed workload-specific tradeoffs, not universal superiority or a causal hardware comparison. The three ARM cycles repeat inference on fixed checkpoints and clouds. Near matched PlantNet accuracy and weaker TomatoWUR stem segmentation further bound the operating range. The evidence supports further embedded integration while keeping energy efficiency, autonomous flight and universal superiority outside the demonstrated claims. Output remains limited to 1,024 sampled labels and, on TomatoWUR, an annotation-prefiltered input; full resolution accuracy and the cost of autonomous scene preparation remain unmeasured.

\appendix
\setcounter{table}{0}
\renewcommand{\thetable}{A.\arabic{table}}
\renewcommand{\theHtable}{A.\arabic{table}}
\section{Additional accuracy and implementation context}
Table~\ref{tab:context} summarizes earlier architectural alternatives. The broader nested AO comparison appears in the main results (Table~\ref{tab:seed42}).
\begin{table}[tbp]\centering\small
\caption{Historical architectural alternatives. LCCA denotes the late cloud context adapter; GLR1 is the geometric logit residual. GLR1 differences use its own paired controls and cannot be added to another campaign's absolute means.}\label{tab:context}\resizebox{\linewidth}{!}{\begin{tabular}{lrrrr}
\toprule
Collection & Multiscale & LCCA & LCCA $-$ Multiscale & GLR1 $-$ Multiscale \\
\midrule
AO & 88.29 & 88.78 & +0.48 & -0.07 \\
Pheno4D & 75.84 & 75.67 & -0.16 & +0.09 \\
ZeaMays & 88.67 & 88.71 & +0.04 & -0.06 \\
\bottomrule
\end{tabular}
}\end{table}

The recorded ternary Multiscale model is also identified as E4 or T3 in historical files; the binary model uses its own frozen configuration. E1 is not the older architecture called T1. Its native XYZ transformation removes 1,152 redundant parameters without retraining. The earlier five process desktop audit preserved logits on 48 real clouds and six numerical stress cases, but did not show a consistent full pipeline speed gain from this algebraic reduction. That result concerns representation simplification, not P4. Successful fixed shape tracing and remaining median related export failures must likewise be kept separate from board execution.

The manuscript source includes the tables and figures needed to compile this article. The accompanying research materials retain the completed factorial summaries, paired contrasts and historical benchmark tables. Explicit cohort filters link displayed baseline means to archived rows. These assets support verification of manuscript arithmetic and figures; they do not independently reproduce training without the source projects and permitted data.

\section{Details of the frozen geometric representation and network}
\label{app:architecture_details}
This specification describes the archived model, without changing its operators or interpreting the implementation optimization as a new architecture. The binary deployment recipe and the ternary AO recipe share these operations but retain their separately recorded class counts, refinement neighborhoods and dropout settings. The source identities are recorded in \texttt{architecture\_source\_audit.json}.

\subsection{Descriptor extraction and normalization}
For each point in the complete retained raw cloud, the loader queries 16 nearest points, centers their coordinates and forms their sample covariance with denominator 15. The query includes the focal point when returned by the neighbor search. Eigenvalues are clamped below at $10^{-12}$ and ordered as $\lambda_1\geq\lambda_2\geq\lambda_3$. The descriptor contains the three coordinates of the eigenvector associated with $\lambda_3$, followed by
\begin{equation}
\begin{split}
&\lambda_3/(\lambda_1+\lambda_2+\lambda_3),\quad
(\lambda_1-\lambda_2)/\lambda_1,\quad
(\lambda_2-\lambda_3)/\lambda_1,\quad
\lambda_3/\lambda_1,\\
&(\lambda_1\lambda_2\lambda_3)^{1/3},\quad
(\lambda_1-\lambda_3)/\lambda_1,\quad
-\sum_{a=1}^{3}\lambda_a\log\lambda_a.
\end{split}
\end{equation}
The last expression uses the clamped eigenvalues themselves, not probabilities obtained by normalizing their sum. All ten descriptor columns are standardized using full cloud means and population standard deviations, with a denominator floor of $10^{-8}$. Sampling subsequently selects the coordinates and corresponding descriptor rows.

Inside the network, the three standardized eigenvector channels are normalized by their vector norm, with a floor of $10^{-6}$. We denote the resulting vector by $u_i$. Because componentwise standardization precedes this operation, $u_i$ must not be interpreted as the unchanged physical surface normal. For each of the seven scalar descriptor channels, the model subtracts its median over the sampled cloud, divides by its interquartile range with a floor of $10^{-6}$, and clips the result to $[-5,5]$. Let $a_i$ denote the first three resulting scalar channels: curvature, linearity and planarity. These transformations retain the historical feature semantics, including the original standardization of the eigenvector channels.

\subsection{Relative messages, fusion and attention}
Let $\mathcal N_k(i)$ be the first $k$ entries of the fixed coordinate neighbor list. Neighbor lists for the two branches are nested and the geometric scale is computed separately within each branch:
\begin{equation}
\begin{split}
s_i^{(k)} &= \max\left(\operatorname{median}_{j\in\mathcal N_k(i)}\|x_j-x_i\|_2,10^{-6}\right),\\
r_{ij}^{(k)} &= \left[(x_j-x_i)/s_i^{(k)},\;
\|x_j-x_i\|_2/s_i^{(k)},\;|u_i^\top u_j|,\;|a_j-a_i|\right].
\end{split}
\end{equation}
Absolute values on $a_j-a_i$ are componentwise, giving eight geometric relation channels in total. The implementation uses \texttt{torch.median}, including its lower middle value for an even number of entries, rather than replacing it with an average of the two central distances. The robust scalar normalization likewise retains the archived median and quantile operations.

Each branch applies an independently learned message MLP to $[h_i,h_j-h_i,r_{ij}^{(k)}]$. Its sequence is linear, ReLU, dropout and linear, with no internal residual connection. A channelwise maximum over neighbors is followed by layer normalization, ReLU and dropout. The $k=8$ and $k=16$ outputs are concatenated and transformed by linear, layer normalization, ReLU and dropout. Both backbone blocks output 32 channels. Their input widths are 13 and 32, respectively, and their message hidden width is 32. Optional third neighborhood branches present in the source are disabled in the recipes reported here.

The two subsequent \texttt{GATConv} layers operate on the fixed $k=16$ coordinate graph, with edges directed from neighbor to focal point. Each uses two heads and 32 channels per head; the first concatenates its heads to 64 channels, and the second averages its heads to 32. ELU and dropout follow each attention layer. A linear map produces the initial class logits. Dropout is inactive during evaluation. The graph is not recomputed from the learned feature space.

\subsection{Residual organ correction}
Let $h_1$, $h_2$ and $h_G$ be the two edge outputs and final attention output. A separate multiscale relative edge block transforms $h_G$ into 64 refinement channels, denoted $h_R$. The refinement neighborhoods are 12 and 32 for the ternary AO reference, and 8 and 32 for the binary deployment reference. The correction head receives
\begin{equation}
c_i=[h_{1,i},h_{2,i},h_{G,i},h_{R,i},
\operatorname{mean}_j h_{G,j},\operatorname{max}_j h_{G,j}],
\end{equation}
with the two cloud summaries broadcast to every point. Thus $c_i$ has 224 channels. A linear map to 96 channels, layer normalization, ReLU, dropout and a final linear map produce two corrections. A learned scalar, clamped to $[0,3]$ during the forward pass, scales these corrections before they are added to the stem and leaf logits. The soil logit in the ternary recipe is unchanged by this branch. The training auxiliary stem head uses the same context with hidden width 32; its evaluation is disabled for the reported inference measurements. Its stored parameters are included in the reported model parameter counts. Optional axial descriptors and auxiliary boundary heads are disabled. Final labels are the argmax over the corrected logits for the 1,024 sampled points.

\section*{Data and code availability}
\ifreviewversion
The repository link is provided in the separate title page to preserve anonymous review.
\else
Code and supporting materials associated with this study are maintained at \url{https://github.com/Ivanrms1/efficient-3d-plant-segmentation}.
\fi
The repository contains source code, configurations and selected figure data; reproducing training also requires the corresponding datasets and trained weights, which are not redistributed there. The accompanying numerical tables document the plotted results and completed ablation contrasts. Public dataset references are given in the bibliography. AO is attributed to the dataset described by \citet{ao2022}. File-level lineage, biological grouping and redistribution permissions are distinct from this bibliographic attribution and remain incompletely documented for the local AO and ZeaMays collections.
\ifreviewversion
\else
\section*{Acknowledgements and funding}
The authors acknowledge institutional support from CINVESTAV and financial assistance from SECIHTI. Jesus Ivan Ruiz-Martinez received financial support from the Cinvestav Guadalajara Unit.
\section*{Declaration of competing interests}
Jesus Ivan Ruiz-Martinez reports financial support from the Cinvestav Guadalajara Unit. Apart from this disclosed support, the authors declared no other known competing financial interests or personal relationships that could have appeared to influence the work in the supplied prior declaration.

\fi
\section*{Declaration of generative AI and AI assisted technologies in manuscript preparation}
OpenAI Codex assisted with evidence inspection, organization of references, manuscript revision and code for figures derived from recorded tables. It also assisted preparation of the vector architecture schematic. No experimental measurements were synthesized for this revision. The authors must review the text, evidence and disclosures before submission.

\bibliographystyle{elsarticle-harv}
\bibliography{references}
\end{document}